\documentclass[10pt]{article}
\usepackage[preprint]{tmlr}
\renewcommand{\headrulewidth}{0pt}

\usepackage{amsmath,amssymb,amsthm}
\usepackage{graphicx}
\usepackage{booktabs}
\usepackage{hyperref}
\usepackage{url}
\usepackage{xcolor}

\newtheorem{remark}{Remark}
\newcommand{\AGREE}{\text{\textsc{Agree}}}
\newcommand{\DISAGREE}{\text{\textsc{Disagree}}}
\newcommand{\SF}{\text{\textsc{Single\_Fallback}}}
\newcommand{\SFs}{\text{\textsc{Single\_Fallback}}\ensuremath{^{*}}}
\newcommand{\INVALID}{\text{\textsc{Invalid}}}
\newcommand{\conf}{\mathrm{conf}}
\newcommand{\valid}{\mathrm{valid}}

\title{Anchor and Check: Deterministic Cross-Prompt,\\Cross-Model Reconciliation for Auditable LLM Decisions}

\author{\name Sivakameswara Reddy Chejarla \email sivakreddyc@gmail.com}

\begin{document}
\maketitle

\begin{abstract}
Pipelines that let a large language model (LLM) make discrete decisions need to know which decisions they may act on unattended, and a record that explains each one. We study a small, fully specified reconciliation contract: commit the target model's greedy answer (the \emph{anchor}), draw one or two greedy \emph{checks} that reach the answer by a different route (an alternate prompt, or a model from another family), and let a pure function return one of five outcomes, of which only \AGREE{} is acted on. On 48,000 model--item evaluations over 6,000 questions, with every method given access to verbalized confidence as a tie-breaker, the main finding concerns where calls come from: pooling another family's samples raises semantic entropy's AUROC by 1.7 to 4.0 points at every budget, and a same-model temperature check is the weakest design. As complete pipelines, the point estimates of the three-call contract and pooled semantic entropy over 3 or 5 samples differ by less than one point, with intervals containing zero. As detectors of the anchor's errors, the contract is ahead of pooled entropy with 3 and 5 samples under both tie-breaker conventions; with 10 samples the point estimates differ by less than one point and the sign depends on the convention. Under the strict \AGREE{} gate it automates 70\% of items and still acts on 16\% of errors. What the contract adds is a fixed committed answer, an explicit outcome and a decision replayable from its record; generation itself was not reproducible on the hosted API.
\end{abstract}

\section{Introduction}

LLMs are now placed inside workflows that must \emph{decide} something: whether an alert is benign, which value a document field holds, which action to take next. The output is consumed by code, not read by a person. Two requirements follow. The pipeline must know which decisions it may act on unattended (\emph{confidence gating}), and when a decision is later questioned the record must show why it was accepted (\emph{auditability}).

Sampling-based methods such as self-consistency~\citep{wang2023selfconsistency}, SelfCheckGPT~\citep{manakul2023selfcheckgpt} and semantic entropy~\citep{kuhn2023semantic,farquhar2024semantic} meet the first requirement by drawing several samples and measuring their agreement. They cost 5 to 10 generations per item and reduce the evidence to a score; which answer is committed is a separate choice (the original semantic-entropy work scores a separately generated answer). We ask how much of their error-detection power a much smaller, fully specified procedure can recover, if the extra calls are spent on samples that make \emph{different} mistakes rather than on more samples of the same distribution.

\begin{figure}[t]
\centering
\includegraphics[width=\linewidth]{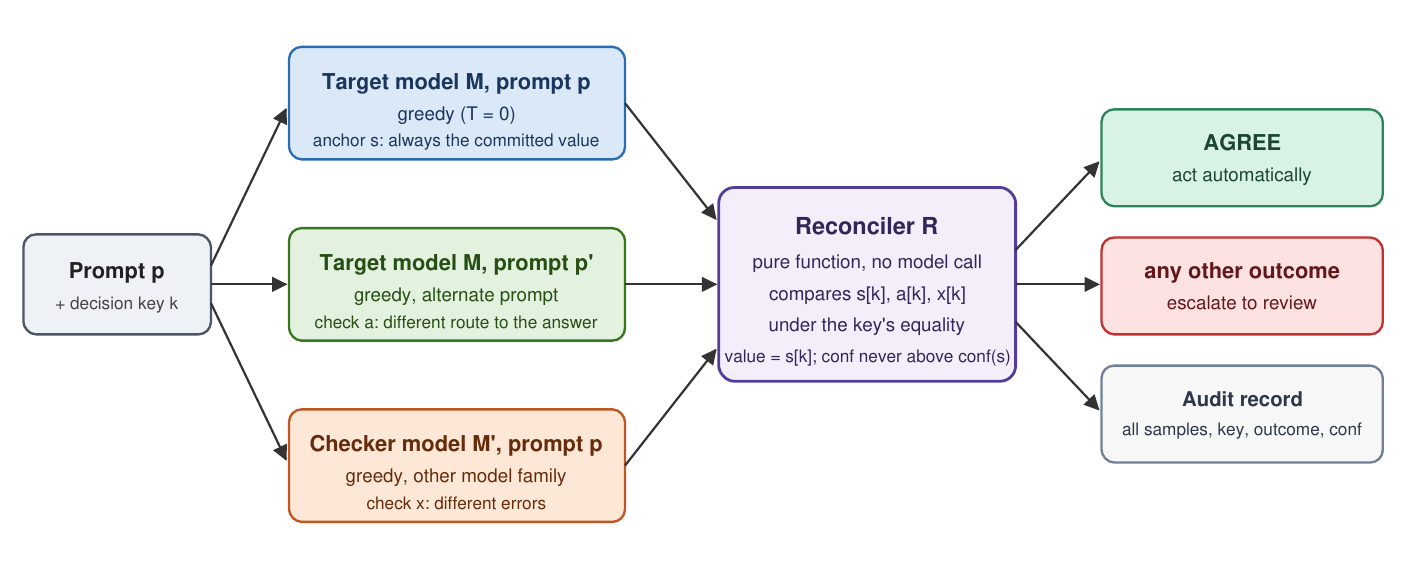}
\caption{Anchor and check. The target model's greedy answer is the anchor and always the committed value. Checks reach the answer by a different route: an alternate prompt, or a model from another family. A pure reconciler compares the declared decision key. Only \AGREE{} is acted on automatically, and the audit record suffices to replay any decision.}
\label{fig:pipeline}
\end{figure}

The procedure has three parts (Figure~\ref{fig:pipeline}). (i)~An \emph{anchor}: the target model's greedy answer, which is always the committed value. (ii)~One or two \emph{checks}, also greedy, chosen to reach the answer by a different route: the same model under an alternate prompt, or a model from a different family. (iii)~A \emph{reconciler}: a pure function that compares a declared decision field across the samples under a per-type equality rule and returns one of five outcomes. Only \AGREE{} is acted on automatically. Both components of the check are known in the hallucination-detection literature~\citep{zhang2023sac3,xue2025verify}; our contribution is a deterministic contract around them and a careful measurement of what it buys.

A natural first design checks the anchor against a sample of the \emph{same} model at temperature 0.8. Under matched scoring that design trails 5-sample semantic entropy by 7 points, because a model's own samples tend to repeat its mistakes (Section~\ref{sec:mechanism}); Appendix~\ref{app:history} records how the design evolved. Changing where the check comes from, not how many checks there are, is what closes the gap. We report the temperature-based design as an ablation. The same lesson applies to the baselines: semantic entropy computed over samples pooled from the target and a peer model improves on single-model semantic entropy at every budget. Compared with these cross-model baselines as complete pipelines, each committing its own answer, the point estimates of the three-call contract and pooled 3- and 5-sample semantic entropy differ by less than one point, with intervals that reach 1.2 points; compared as detectors of the anchor's errors with the same tie-breaker, it is 3.9 points above pooled 3-sample entropy, 1.5 above 5 samples and $+0.1$ points $[-0.6, +0.7]$ from 10 samples (sample counts are the baseline's; the contract uses three calls). Most of that detector lead comes from where the three answers are drawn, not from the scoring rule: entropy over the same three greedy answers is only 0.6 points lower. We do not claim that anchor preservation or replayability is unique to the contract. Semantic entropy can score a separately generated answer~\citep{kuhn2023semantic}, SAC$^3$ assesses the target model's own response~\citep{zhang2023sac3}, and any baseline can store its samples and threshold for replay. What the contract fixes is the combination: a committed answer that never changes with sampling, an explicit five-way outcome in place of a threshold on a score, a confidence bounded by the anchor's, and a specification small enough to be re-implemented from this paper.

Our contributions are:
\begin{itemize}
\item \textbf{A reconciliation contract} (Section~\ref{sec:method}), fully specified: an anchor and $m$ ordered checks, a declared decision key, per-type comparison relations with stated tolerances, five exhaustive outcomes, a confidence that never exceeds the anchor's, and a versioned audit record. It makes no model call and has no randomness, so any decision can be replayed from its record.
\item \textbf{A matched-score evaluation} (Sections~\ref{sec:setup}--\ref{sec:results}) on 48,000 model--item evaluations in which every method has access to verbalized confidence as a tie-breaker, with native-signal results reported alongside, cross-model baselines at equal call budget (pooled semantic entropy and a SAC$^3$-style score), and entailment-based clustering for the short-text tasks. Pooling a second model's samples helps semantic entropy more than adding samples of the same model. As complete pipelines the three-call contract and pooled semantic entropy at 3 and 5 samples differ by less than one point; as detectors of the anchor's errors under one tie-breaker, the contract is 3.9 and 1.5 points above them, and its point estimate is within one point of 10 samples. A decomposition separates the choice of calls (most of the lead) from the scoring rule (0.6 points). A fourth check (the peer under the alternate prompt) adds 1.7 points.
\item \textbf{A measured mechanism} (Section~\ref{sec:mechanism}): per task, the accuracy of each check type and how often it repeats a wrong anchor.
\item \textbf{Confirmation} on a fresh task (CommonsenseQA) run after every design choice was fixed, including the cross-model baselines, and a post hoc held-out half as a robustness check.
\item \textbf{Operational measurements}: two gating policies with risk--coverage curves, gate stability across repeat runs, single-call latency, and cost at the provider's public list prices.
\end{itemize}

\section{Related Work}

\paragraph{Sampling-based uncertainty.} Self-consistency~\citep{wang2023selfconsistency} takes a majority over sampled chain-of-thought answers. SelfCheckGPT~\citep{manakul2023selfcheckgpt} scores a response by its consistency with stochastic re-samples. Semantic entropy~\citep{kuhn2023semantic,farquhar2024semantic} clusters samples by meaning and computes entropy over the clusters; we use its discrete form. The original work scores a separately generated answer with an independently sampled uncertainty set, so it does not require committing a voted answer; our baselines commit the plurality answer, and Section~\ref{sec:crossmodel} also grades them on the anchor. \citet{lyu2024calibrating} calibrate confidence from sample consistency, and BSDetector~\citep{chen2024bsdetector} combines consistency with self-reflection. All of these draw several samples from one model under one prompt.

\paragraph{Perturbed prompts and other models.} SAC$^3$~\citep{zhang2023sac3} is the closest prior work: it adds semantically equivalent question rephrasings and cross-model responses to self-consistency, and shows that each detects errors that self-consistency misses. \citet{xue2025verify} find that self-consistency methods are close to a black-box oracle and that a second verifier model raises that ceiling; they call the verifier only when self-consistency is uncertain; we evaluate both that ordering and an author-designed reverse (Section~\ref{sec:crossmodel}). Option order is a known source of variation in multiple-choice answers~\citep{zheng2024mcq}, which our alternate prompt uses deliberately. Our checks are therefore not new. What we add is a fixed, greedy, minimal configuration (one anchor, at most two checks), a deterministic contract with an audit record around it, matched-score comparisons against single-model and cross-model semantic entropy and a SAC$^3$-style score on the same sample pool, and measurements of cost, latency and stability.

\paragraph{Verbalized confidence.} Models can state a confidence that carries signal~\citep{kadavath2022know,tian2023justask}, though it is often overconfident~\citep{xiong2024express}. We use it only as a tie-breaker and give every method access to it. Each method aggregates it in its own way (Section~\ref{sec:score}); only Table~\ref{tab:controlled} gives every method the identical tie-breaker scalar.

\paragraph{Deterministic guardrails.} \citet{wahi2026judge} and \citet{nam2026integrity} place deterministic acceptance checks around LLM judges and pipeline stages, and \citet{srinivasan2026runtime} catalogs proposer/verifier/commit patterns for agents. Our contract works inside a single step and produces exactly the kind of deterministic evidence such stage gates consume. \citet{ali2026shallow} show that temperature-induced variation carries per-question but little cross-question structure, which is consistent with our finding that temperature is a weak source of checks.

\section{Method}\label{sec:method}

\subsection{Anchor and checks}\label{sec:samples}
Given a prompt $p$, a decision key $k$ and a target model $M$, the \emph{anchor} is $s = M(p)$ under greedy decoding (temperature 0).\footnote{Our harness also sends top-$p$ = 0.7; at temperature 0 it has no effect, so the anchor is plain greedy decoding.} A \emph{check} is any other sample of the same item. We study three kinds, all greedy:
\begin{itemize}
\item \textbf{alt}: $M(p')$, the same model under an alternate prompt $p'$ that keeps the task and output format but changes the route to the answer. The templates were written once and not tuned (Appendix~\ref{app:prompts}): multiple-choice options are listed in reverse order and the chosen letter is mapped back; math problems ask for the equations first; short-answer questions ask the model to name the fact it relies on.
\item \textbf{peer}: $M'(p)$, a model from a different family in the same size class (fixed in advance; Section~\ref{sec:setup}).
\item \textbf{cheap}: $M''(p)$, the other-family model with the lowest measured cost per call.
\end{itemize}
As an ablation we also use the check we started from, \textbf{creative}: $M(p)$ at temperature 0.8, top-$p$ 1.0.

\subsection{Decision key and equality}\label{sec:eq}
Downstream code acts on a specific field, so the reconciler compares only $s[k]$ against each check's $c[k]$ and ignores the rest of the output. The comparison relation $\approx_k$ is chosen per key type (Table~\ref{tab:eq}). Numeric values are parsed after stripping currency symbols and thousands separators and compared with a relative tolerance of $10^{-6}$ (absolute below 1). For short text the reconciler uses normalized exact match; the gold alias lists of TriviaQA and NQ are used only to grade correctness, never inside the contract. Two of the relations are not equivalences. The date relation is a \emph{compatibility} relation: a missing month or day is a wildcard, so \texttt{2000} is compatible with both \texttt{2000-01-01} and \texttt{2000-02-02}, which are not compatible with each other. Numeric tolerance is likewise not transitive in general (two values each within tolerance of a third need not be within tolerance of each other), although at $10^{-6}$ this never arose in our data. The contract never needs transitivity, because it compares each check with the anchor only (a star, not a clique). Where a method clusters samples (semantic entropy), we cluster greedily in sample order: a sample joins the first existing cluster whose representative it is compatible with, otherwise it opens a new one. Under a non-transitive relation the clusters, and so the entropy, can depend on sample order; we use the order in which samples were drawn, the same for every method and every key type. The wildcard rule also means that the comparator discards precision: an anchor of \texttt{2000-01-01} and a check of \texttt{2000} reconcile to \AGREE{} even if only one of them matches what the source states. The structured and date relations are specified and unit-tested but not evaluated: none of the tasks in this paper uses a structured or date decision key.

\begin{table}[t]
\centering\small
\caption{Comparison relations used by the reconciler, by key type. The categorical, short-text and structured relations are equivalences; the date relation is a compatibility relation and numeric tolerance is not transitive in general (Section~\ref{sec:eq}).}
\label{tab:eq}
\begin{tabular}{lll}
\toprule
Key type & Normalization & Equality \\
\midrule
Numeric & parse; strip units and thousands separators & numeric equality within tolerance \\
Categorical & case-fold, strip whitespace & exact \\
Short text & case-fold, strip punctuation and articles & exact \\
Structured & canonical JSON (sorted keys, normalized scalars) & exact \\
Date & parse ISO or prose; missing month/day is a wildcard & component-wise match \\
\bottomrule
\end{tabular}
\end{table}

\subsection{The reconciliation contract}\label{sec:contract}
\begin{figure}[t]
\centering
\includegraphics[width=\linewidth]{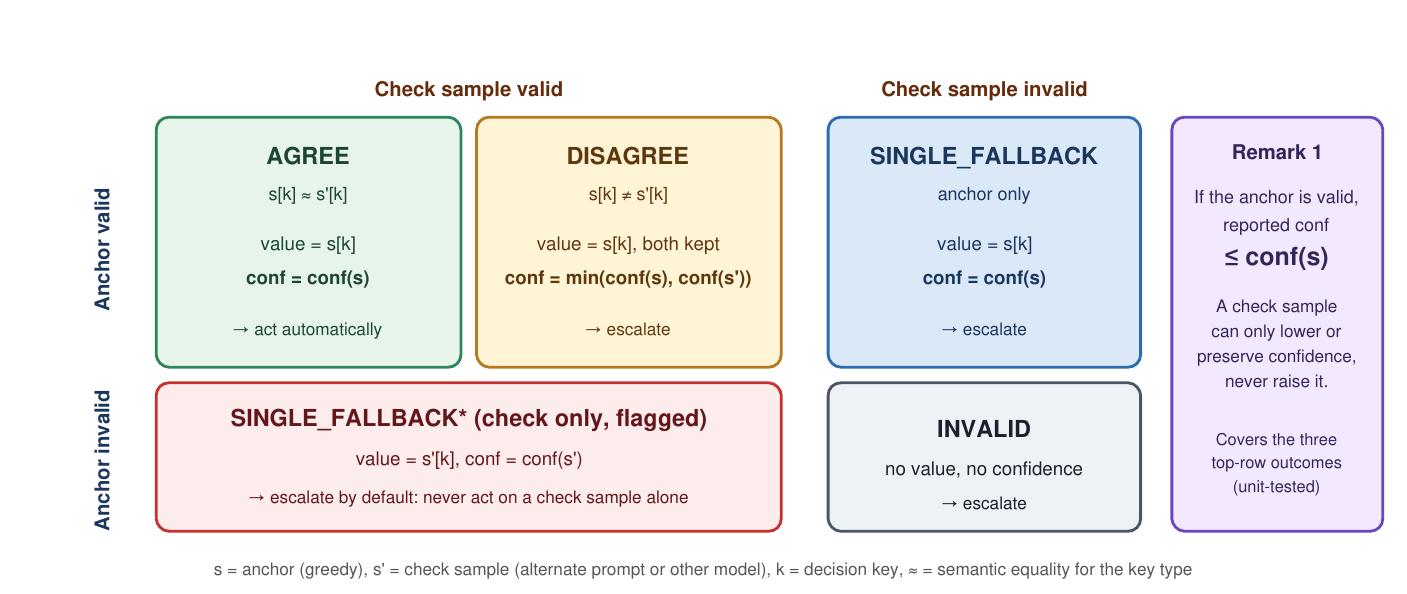}
\caption{Outcomes for one check ($m=1$) and the confidence each reports; $s$ is the anchor and $s'$ the check. Only \AGREE{} is acted on automatically; every other outcome is escalated.}
\label{fig:outcomes}
\end{figure}

Let $\valid(x)$ hold if $x$ parses and $x[k]$ is present and well-typed. The checks are given in a declared order $c_1,\dots,c_m$ (in our runs: alt, then peer). $\conf(x)$ is the verbalized confidence parsed from the sample's \texttt{Confidence:} line, clipped to $[0,100]$ and scaled to $[0,1]$; a missing or unparseable line gives $0.5$. The reconciler $R$ returns (outcome, value, confidence):
\begin{itemize}
\item \INVALID{} $(\bot,\bot)$ if no sample is valid;
\item \SFs{} $(c_j[k], \conf(c_j))$, flagged, if $s$ is invalid and $j$ is the smallest index with $\valid(c_j)$. Valid checks may disagree with each other; the record keeps all of them, and because \SFs{} is never acted on, the choice of $j$ affects only the recorded value;
\item \SF{} $(s[k], \conf(s))$ if $s$ is valid, at least one check is invalid, and every valid check satisfies $c_j[k]\approx_k s[k]$;
\item \AGREE{} $(s[k], \conf(s))$ if all $m{+}1$ samples are valid and $c_j[k]\approx_k s[k]$ for every $j$;
\item \DISAGREE{} $(s[k], \min_{x\ \text{valid}}\conf(x))$ otherwise.
\end{itemize}
The cases are exhaustive and mutually exclusive, so $R$ is total; it makes no model call and uses no randomness, so the same samples always give the same outcome, value and confidence. The work is linear in $m$ plus the cost of parsing and comparing $m{+}1$ fields; for $m\le 2$ this is microseconds. The deployment policy acts automatically on \AGREE{} only and escalates every other outcome; \SFs{} is flagged because it rests on a check alone. Figure~\ref{fig:outcomes} lays out the outcomes for $m=1$. The audit record is a versioned JSON object with: the record schema version; the equality-rule version; the decision key and key type; every raw sample with its source (model identifier, prompt identifier, decoding settings, timestamp, token usage); the parsed value and confidence of each sample; the outcome; the committed value and confidence; and the policy action taken. Replaying $R$ over the stored samples reproduces the decision exactly.

\begin{remark}\label{rem:bound}
If the anchor is valid, the reported confidence is at most $\conf(s)$, because every reachable outcome reports either $\conf(s)$ or a minimum that includes it. A check can lower confidence but never raise it. This is an algebraic property of $R$, not a calibration or correctness guarantee. The committed value is the anchor's whenever the anchor is valid; when it is invalid the \SFs{} value is recorded but never acted on, and we grade the item as wrong. Under this convention the contract's accuracy equals greedy accuracy exactly, by construction; we treat this as a design property, not a finding.
\end{remark}

\subsection{Scores used for evaluation}\label{sec:score}
To rank items for AUROC we need a scalar. For one check we order outcomes \AGREE{} $>$ \SF{} $>$ \DISAGREE{} $=$ \SFs{} $>$ \INVALID{} and break ties with the contract confidence (the anchor's on \AGREE{} and \SF, the minimum on \DISAGREE). For two checks the native score is the number of checks that agree with the anchor ($-1$ if the anchor is invalid), with ties broken by the anchor's confidence. The tie-breaker is added with weight $10^{-3}$, so it only orders items whose native scores are equal. Semantic entropy's tie-breaker is the mean verbalized confidence of the samples in its plurality class. These scalars are not identical across methods, so Section~\ref{sec:crossmodel} also reports a controlled comparison in which every method uses the anchor's confidence as the tie-breaker and predicts the anchor's correctness.

\section{Experimental Setup}\label{sec:setup}

\paragraph{Tasks.} Six tasks with a natural decision key, 1,000 items each, sampled with a fixed seed, so the main grid is 48,000 model--item evaluations over 6,000 distinct questions: GSM8K and SVAMP (final number, numeric equality), MMLU and ARC-Challenge (option letter, categorical), TriviaQA and NQ-Open (short answer, short text). A seventh task, CommonsenseQA~\citep{talmor2019commonsenseqa} (validation split, 1,000 items, 5 options, MIT license), was added only after the method and checker assignments were fixed, and is used solely for confirmation.

\paragraph{Models.} Eight open-weight models from 4 families, two sizes each, served through one hosted inference API in a single region: Ministral 3 8B and Mistral Large 3; Qwen3 32B and Qwen3 235B A22B; gpt-oss-20b and gpt-oss-120b; GLM 4.7-flash and GLM 5. The gpt-oss models run with \texttt{reasoning\_effort=low} and 512 extra output tokens per call, because their reasoning counts against the output budget. After one format-repair call (the same model restating its own answer), 0.18\% of anchors and 0.10\% of alternate-prompt checks remain unparseable; per-model rates are in Appendix~\ref{app:permodel}.

\paragraph{Checker assignment.} Fixed before any result was computed. \emph{peer} pairs models of the same size class from different families: Ministral 8B $\leftrightarrow$ GLM 4.7-flash, Qwen3 32B $\leftrightarrow$ gpt-oss-20b, Mistral Large 3 $\leftrightarrow$ GLM 5, Qwen3 235B $\leftrightarrow$ gpt-oss-120b. \emph{cheap} is GLM 4.7-flash for the six non-GLM targets and Ministral 8B for the two GLM targets. All 8 models answered the same items under the same prompt, so a checker's answer is its own anchor for that item; a repeat run (Section~\ref{sec:stability}) shows what a fresh checker call changes.

\paragraph{Methods.} Every method draws from one shared pool per item: the anchor, 10 samples at $T{=}0.8$, the alternate-prompt sample, and the other models' anchors.
\begin{itemize}
\item \textbf{Contract, 2 calls}: anchor + alt, anchor + peer, anchor + cheap; ablation: anchor + creative.
\item \textbf{Contract, 3 calls}: anchor + alt + peer, anchor + alt + cheap.
\item \textbf{Baselines}: greedy with verbalized confidence (1 call); discrete semantic entropy (SE) over $N\in\{3,5,10\}$ samples at $T{=}0.8$, committing the plurality answer and scored by negative entropy. SE also has access to verbalized confidence: its tie-breaker is the mean verbalized confidence of the samples in the plurality class. SE clusters samples with the same equality relation as the reconciler (Table~\ref{tab:eq}); for the two short-text tasks we also report both SE and the contract with an entailment-based relation (Section~\ref{sec:entail}).
\item \textbf{Cross-model baselines at equal budget}: \emph{pooled SE-$N$}, semantic entropy over $T{=}0.8$ samples split between the target and its peer (1+1, 2+1, 3+2, 5+5; ties go to the target's answer); \emph{pooled greedy SE}, semantic entropy over exactly the greedy answers the contract uses (anchor and peer; anchor, alt and peer); and a \emph{SAC$^3$-style} score~\citep{zhang2023sac3}: the number of checks that agree with the anchor among alt, peer and the peer under the alternate prompt (4 calls), with equal weights. The SAC$^3$-style score is the contract with a third check added, so it also commits the anchor.
\item \textbf{Hybrids}: in the ordering of \citet{xue2025verify}, self-consistency first ($n\in\{3,5\}$ samples of the target), with the anchor and the peer called only when the samples are not unanimous (specified in Section~\ref{sec:crossmodel}); and an author-designed reverse, anchor and peer first with SE-5 only on \DISAGREE.
\end{itemize}

\paragraph{Metrics.} The headline is error-detection AUROC averaged over the 48 task-model cells, so that confidence tracking task difficulty does not inflate it. Differences between methods carry 95\% paired bootstrap intervals with 1,000 resamples. Each resample draws item identifiers once per task and applies the draw to all 8 models and to both methods, because the 8 cells of a task share items and each peer pair's calls serve two cells; the 48 cells are therefore not treated as independent. Each method is graded on the answer it commits: the anchor for the contract, the plurality answer for SE; Section~\ref{sec:crossmodel} adds anchor-graded and controlled comparisons. We did not pre-specify an equivalence margin, so we report effect estimates with their 95\% intervals and make no equivalence claim; an interval that contains zero means the data do not resolve the sign of the difference, not that the methods are equal. The same item-clustered bootstrap gives every interval in the paper, including the gating results. Cost uses token counts reported by the API for every call (including repair calls) and the provider's public on-demand list prices for the region used (price list of 2026-11-11). We also report gating, stability and latency (Section~\ref{sec:ops}).

\section{Results}\label{sec:results}

\subsection{Error detection and cost}\label{sec:main}
\begin{figure}[t]
\centering
\includegraphics[width=\linewidth]{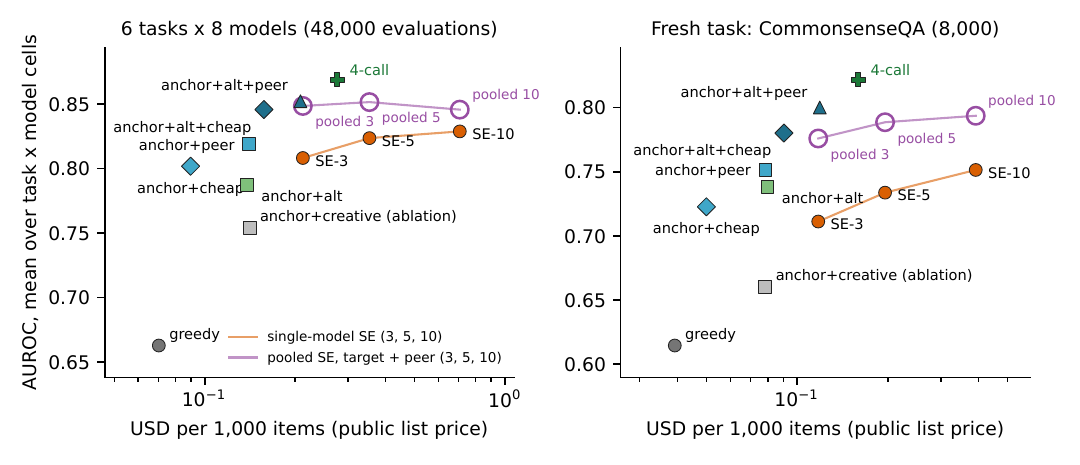}
\caption{Error-detection AUROC (mean over task-model cells) against cost per 1,000 items. Left: main grid. Right: the fresh CommonsenseQA task. Semantic-entropy points are joined by a line.}
\label{fig:cost}
\end{figure}

Table~\ref{tab:main} gives the main grid against single-model baselines. With one cross-family check, the contract reaches 0.819, against 0.824 for SE-5; the paired difference is $-0.5$ points $[-1.3, +0.4]$, an interval that leaves the sign unresolved. It costs 40\% of SE-5 (USD 0.141 against 0.354 per 1,000 items) and beats SE-3 by 1.1 points $[+0.2, +1.9]$ at two-thirds of the calls. The cheap checker gives 0.802 at a quarter of SE-5's cost and is 0.6 points below SE-3 $[-1.7, +0.5]$. The alternate prompt alone is weaker (0.787), but it adds information the peer does not: anchor + alt + peer reaches \textbf{0.852}, 2.9 points above SE-5 $[+2.1, +3.7]$ and 2.3 points above SE-10 $[+1.6, +3.2]$, at 29\% of SE-10's cost. Figure~\ref{fig:cost} plots AUROC against cost. These comparisons are against baselines that use one model; Section~\ref{sec:crossmodel} gives baselines that use two.

The temperature-based check is the weakest two-call design (0.754, 6.5 points below the peer check at the same cost). Section~\ref{sec:mechanism} measures why.

\begin{table}[t]
\centering\small
\caption{Main grid: 6 tasks $\times$ 8 models, 48,000 model--item evaluations over 6,000 questions. AUROC is the mean over 48 task-model cells; ``native'' drops the verbalized-confidence tie-breaker for every method. Accuracy of every contract row is the greedy accuracy by construction. Cost: USD per 1,000 items at public list prices.}
\label{tab:main}
\begin{tabular}{lccccc}
\toprule
Method & Calls & Accuracy & AUROC & AUROC (native) & USD / 1k items \\
\midrule
Greedy + verbalized confidence & 1 & 0.764 & 0.663 & 0.515 & 0.070 \\
\midrule
Anchor + creative (ablation) & 2 & 0.764 & 0.754 & 0.679 & 0.141 \\
Anchor + alt & 2 & 0.764 & 0.787 & 0.728 & 0.138 \\
Anchor + cheap & 2 & 0.764 & 0.802 & 0.761 & \textbf{0.090} \\
\textbf{Anchor + peer} & 2 & 0.764 & \textbf{0.819} & \textbf{0.776} & 0.141 \\
\midrule
Anchor + alt + cheap & 3 & 0.764 & 0.846 & 0.822 & 0.158 \\
\textbf{Anchor + alt + peer} & 3 & 0.764 & \textbf{0.852} & \textbf{0.826} & 0.208 \\
\midrule
Semantic entropy, 3 samples & 3 & 0.763 & 0.808 & 0.733 & 0.212 \\
Semantic entropy, 5 samples & 5 & 0.769 & 0.824 & 0.765 & 0.354 \\
Semantic entropy, 10 samples & 10 & 0.775 & 0.829 & 0.784 & 0.708 \\
\bottomrule
\end{tabular}
\end{table}

\paragraph{Matched scoring.} The tie-breaker helps every method, and helps semantic entropy as much as ours (SE-5 rises from 0.765 to 0.824). It does change one ordering: on the native signals alone, anchor + peer is 1.1 points above SE-5 (0.776 against 0.765), and with the tie-breaker it is 0.5 below (Table~\ref{tab:main}, native column); anchor + alt + peer is above SE-10 in both views (by 4.2 and 2.3 points). Under the alternative convention of grading SE on the anchor rather than on its own plurality answer, SE-3 and SE-5 score lower (0.786 and 0.814) and SE-10 slightly higher (0.835), still 1.7 points below the three-call contract. This convention also explains why SE-10 scores below SE-5 on three models in Table~\ref{tab:permodelfull} (Ministral 8B, Mistral Large 3, gpt-oss-20b): graded on the anchor, SE-10 is above SE-5 on all three (for example 0.849 against 0.821 on Ministral 8B). The dip comes from committing the plurality answer, whose errors with ten samples are harder for the entropy to rank, not from the clustering (Section~\ref{sec:entail}).

\paragraph{Accuracy.} The contract commits the anchor, so its accuracy is greedy accuracy (0.764). Semantic entropy's vote adds 0.5 points at 5 samples and 1.1 at 10. Pooled semantic entropy, whose vote spans two models, reaches 0.767 to 0.791. The extra accuracy does not have to cost 2.5--5$\times$: a vote over the contract's own three greedy answers reaches 0.779 at the contract's cost (Table~\ref{tab:crossmodel}, pooled greedy SE-3). What the contract gives up by committing the anchor is that 1.5 points of accuracy, in exchange for a committed answer that no sampling can move.

\subsection{Cross-model baselines at equal budget}\label{sec:crossmodel}
The comparisons above set a two-model contract against one-model baselines. Table~\ref{tab:crossmodel} removes that asymmetry by giving the baselines the same peer. Pooling helps semantic entropy at every budget: pooled SE-3 is 4.0 points above SE-3 $[+3.4, +4.7]$, pooled SE-5 2.8 points above SE-5 $[+2.1, +3.5]$, and pooled SE-10 1.7 points above SE-10 $[+1.0, +2.4]$. Pooled SE-3 (0.849) is above single-model SE-10 (0.829) by 2.0 points $[+1.2, +2.7]$. This is the direct test of the claim that the source of the calls matters more than their number, and it holds for the baseline as well as for the contract.

\paragraph{As complete pipelines.} Graded on the answer each method commits (Table~\ref{tab:crossmodel}), the three-call contract is not ahead of pooled semantic entropy: $+0.4$ points from pooled SE-3 $[-0.5, +1.2]$ and $+0.1$ from pooled SE-5 $[-0.8, +1.0]$. Both intervals contain zero; they are consistent with differences of up to 1.2 points in either direction, and since we did not pre-specify an equivalence margin we make no equivalence claim. Pooled semantic entropy also commits a more accurate voted answer (0.767--0.791 against 0.764). At 2 calls the contract loses: semantic entropy over the anchor and the peer, the same two answers, scores 0.9 points higher $[-1.3, -0.5]$, and pooled SE-2 over one $T{=}0.8$ sample per model scores 0.842.

\paragraph{As detectors of the anchor's errors.} The pipelines commit different answers, so their AUROCs measure different things. Grading every method on the anchor's correctness (Table~\ref{tab:anchorgraded}, each method keeping its own tie-breaker) changes the picture for the sampled pools: pooled SE-2 at $T{=}0.8$ falls from 0.842 to 0.785, because many of the errors it detected were its own sampled answer's, not the anchor's; pooled SE-3 falls to 0.819, 3.4 points below the three-call contract $[+2.6, +4.3]$. The contract is 0.8 points above pooled SE-5 $[+0.1, +1.6]$, and 0.5 below pooled SE-10 $[-1.2, +0.3]$ and 0.2 below a vote over its own three greedy answers $[-0.6, +0.1]$. The controlled comparison in Table~\ref{tab:controlled} goes one step further and also gives every method the same tie-breaker scalar, the anchor's verbalized confidence. The table counts calls two ways: the \emph{extra} verification calls, and the \emph{total} including the anchor's greedy call, which every sampled baseline in this table also needs because it supplies the label target and the tie-breaker (so pooled SE-3 is four calls in total against the contract's three). Under this control, anchor + peer and semantic entropy over the same two greedy answers are identical (they induce the same partition), and the three-call contract is $+3.9$ points $[+3.3, +4.6]$ above pooled SE-3, $+1.5$ $[+0.8, +2.2]$ above pooled SE-5 and $+0.1$ $[-0.6, +0.7]$ from pooled SE-10. Both views agree that the contract is ahead of pooled entropy over 3 and 5 samples; at 10 samples the sign depends on the tie-breaker, so we report only that the point estimates differ by less than one point there ($+0.1$ $[-0.6, +0.7]$ under the anchor's tie-breaker, $-0.5$ $[-1.2, +0.3]$ under each method's own).

\paragraph{Where the detector lead comes from.} Two rows of Table~\ref{tab:controlled} separate the choice of calls from the scoring rule. Entropy over the \emph{same} three greedy answers (anchor, alternate prompt, peer) is only 0.6 points below the contract $[+0.5, +0.7]$: that is the whole value of the agreement count as a scoring rule. The remaining 3.3 points of the 3.9-point lead over pooled sampled SE-3 come from which three answers are collected. The same-model rows make the other cut: with the contract's own scoring rule (agreement with the anchor) but checks drawn as temperature samples of the target, two checks reach 0.788, four reach 0.814 and nine reach 0.836, all below the three-call contract. Diverse checks, not the rule, carry most of the effect, which is also what Section~\ref{sec:mechanism} measures. The 0.6 points the rule does add have a simple cause: among the 14.5\% of items where the three greedy answers split two to one with the anchor in the majority, the anchor is wrong 45\% of the time, and among the 3.3\% where the two checks agree against it, 79\%. Entropy over three answers assigns both configurations the same value; the agreement count assigns 1 and 0. The 2-call difference between the two gradings is entirely the tie-breaker: entropy averages both models' confidences, while the contract reports the anchor's on \AGREE{} and the minimum on \DISAGREE{} (Remark~\ref{rem:bound}); that bound costs 0.9 points of pipeline AUROC.

\paragraph{The four-call configuration.} Adding the peer's answer under the alternate prompt as a third check ($m{=}3$) gives the best detector in the paper, 0.869, 1.7 points above the three-call contract $[+1.4, +2.0]$ at USD 0.276 per 1,000 items. It is the same contract with one more check. We label it SAC$^3$-style because it combines a prompt perturbation with a cross-model check as SAC$^3$ does~\citep{zhang2023sac3}; published SAC$^3$ checks semantically equivalent question rephrasings and combines the self- and cross-model scores with a weight parameter whose default is 1. We use no rephrasings (the alternate prompt takes their place) and count every check equally.

\paragraph{Hybrids.} Our adaptation of the ordering of \citet{xue2025verify} runs self-consistency first. Per item: (1)~draw $n$ samples of the target at $T{=}0.8$ ($n$ calls); (2)~if they are unanimous, commit their answer with score $1 + \epsilon\bar c$, where $\bar c$ is the samples' mean verbalized confidence; (3)~otherwise call the target greedily (the anchor) and the peer greedily (2 more calls), commit the anchor, and score the pair by its outcome tier minus 3 (\AGREE{} gives 0, below every unanimous item), with no tie-breaker. Calls are $n$ on the unanimous branch and $n{+}2$ otherwise. With $n{=}3$ the samples split on 23.0\% of items, for 3.46 calls and USD 0.234 per 1,000 items on average; with $n{=}5$ on 27.5\%, for 5.55 calls and USD 0.380. Pipeline AUROC is 0.805 and 0.830. For an anchor-graded score we label every item by the pool's anchor, which the pipeline itself generates only on the split branch; this gives 0.795 and 0.820, below the three-call contract by 5.7 $[+5.0, +6.6]$ and 3.2 $[+2.5, +4.0]$ points, at higher cost. Our author-designed reverse (anchor and peer first, SE-5 only on \DISAGREE) scores 0.807 at 3.2 calls, below anchor + peer alone (0.819) despite the extra calls. The reason is the committed answer: on the 24\% of items that escalate, the hybrid commits SE-5's plurality answer, which changes the committed answer on 2.7\% of all items (1.9\% from wrong to right, 0.8\% the reverse). The corrected items become correct items with low scores, which lowers the pipeline AUROC even though the detector improves (anchor-graded: 0.832).

\begin{table}[t]
\centering\small
\caption{Cross-model baselines at equal call budget, complete pipelines, main grid (mean over 48 cells). Each method is graded on the answer it commits. ``Calls'': calls the pipeline makes. ``Diff.'': contract minus baseline, paired 95\% interval from the clustered bootstrap; the reference contract is anchor + peer for the 2-call block and anchor + alt + peer otherwise. Cost in USD per 1,000 items at public list prices, repair calls included; adaptive methods are averaged over the calls each item actually triggers. Table~\ref{tab:anchorgraded} grades the same scores against the anchor's correctness.}
\label{tab:crossmodel}
\resizebox{\linewidth}{!}{\begin{tabular}{lccccc}
\toprule
Method & Calls & USD / 1k & Accuracy & AUROC & Contract $-$ baseline \\
\midrule
\textbf{Anchor + peer} (contract) & 2 & 0.141 & 0.764 & 0.819 & \\
Pooled greedy SE-2 (anchor, peer) & 2 & 0.141 & 0.765 & 0.828 & $-0.9$ $[-1.3, -0.5]$ \\
Pooled SE-2 ($T{=}0.8$, 1 target + 1 peer) & 2 & 0.142 & 0.752 & 0.842 & $-2.4$ $[-3.1, -1.5]$ \\
\midrule
\textbf{Anchor + alt + peer} (contract) & 3 & 0.208 & 0.764 & 0.852 & \\
Pooled greedy SE-3 (anchor, alt, peer) & 3 & 0.208 & 0.779 & 0.838 & $+1.4$ $[+0.8, +2.1]$ \\
Pooled SE-3 (2 target + 1 peer) & 3 & 0.212 & 0.767 & 0.849 & $+0.4$ $[-0.5, +1.2]$ \\
Pooled SE-5 (3 target + 2 peer) & 5 & 0.354 & 0.780 & 0.852 & $+0.1$ $[-0.8, +1.0]$ \\
Pooled SE-10 (5 target + 5 peer) & 10 & 0.708 & 0.791 & 0.846 & $+0.7$ $[-0.3, +1.8]$ \\
4-call configuration (alt, peer, peer-alt) & 4 & 0.276 & 0.764 & 0.869 & $-1.7$ $[-2.0, -1.4]$ \\
\midrule
Hybrid, \citet{xue2025verify} order: SE-3, then anchor + peer if split & 3.46 & 0.234 & 0.768 & 0.805 & \\
Hybrid, \citet{xue2025verify} order: SE-5, then anchor + peer if split & 5.55 & 0.380 & 0.766 & 0.830 & \\
Hybrid, reverse (ours): anchor + peer, SE-5 on \DISAGREE & 3.2 & 0.198 & 0.775 & 0.807 & \\
\bottomrule
\end{tabular}}
\end{table}

\begin{table}[t]
\centering\small
\caption{The scores of Table~\ref{tab:crossmodel} as detectors of the anchor's errors: every score is graded against the anchor's correctness, each method keeping its own tie-breaker. ``Total calls'' includes the anchor's greedy call, which a sampled baseline does not make itself but needs here because the anchor is the label target; this is why pooled SE-3 counts 3 calls in Table~\ref{tab:crossmodel} and 4 here, as in Table~\ref{tab:controlled}. ``Diff.'': reference contract minus baseline, paired 95\% interval.}
\label{tab:anchorgraded}
\begin{tabular}{lccc}
\toprule
Method & Total calls & AUROC & Contract $-$ baseline \\
\midrule
\textbf{Anchor + peer} (contract) & 2 & 0.819 & \\
Pooled greedy SE-2 (anchor, peer) & 2 & 0.831 & $-1.3$ $[-1.6, -0.9]$ \\
Pooled SE-2 ($T{=}0.8$, 1 target + 1 peer) & 3 & 0.785 & $+3.4$ $[+2.5, +4.4]$ \\
\midrule
\textbf{Anchor + alt + peer} (contract) & 3 & 0.852 & \\
Pooled greedy SE-3 (anchor, alt, peer) & 3 & 0.854 & $-0.2$ $[-0.6, +0.1]$ \\
Pooled SE-3 (2 target + 1 peer) & 4 & 0.819 & $+3.4$ $[+2.6, +4.3]$ \\
Pooled SE-5 (3 target + 2 peer) & 6 & 0.844 & $+0.8$ $[+0.1, +1.6]$ \\
Pooled SE-10 (5 target + 5 peer) & 11 & 0.857 & $-0.5$ $[-1.2, +0.3]$ \\
4-call configuration (alt, peer, peer-alt) & 4 & 0.869 & $-1.7$ $[-2.0, -1.4]$ \\
\midrule
Hybrid, \citet{xue2025verify} order, SE-3 & 4.23 & 0.795 & $+5.7$ $[+5.0, +6.6]$ \\
Hybrid, \citet{xue2025verify} order, SE-5 & 6.27 & 0.820 & $+3.2$ $[+2.5, +4.0]$ \\
Hybrid, reverse (ours) & 3.2 & 0.832 & $+2.1$ $[+1.5, +2.6]$ \\
\bottomrule
\end{tabular}
\end{table}

\begin{table}[t]
\centering\small
\caption{Controlled detector comparison, main grid: every method predicts the anchor's correctness and uses the anchor's verbalized confidence as its tie-breaker. ``Extra'': verification calls beyond the anchor; ``Total'': including the anchor's greedy call, which every row needs for the label target and the tie-breaker. Mean AUROC over 48 cells. ``Diff.'' is the block's reference contract minus the baseline, paired 95\% interval: anchor + peer in the upper block, anchor + alt + peer in the lower one. ``Same-model, agreement'': the contract's scoring rule applied to $k$ temperature-0.8 samples of the target in place of the alternate prompt and the peer.}
\label{tab:controlled}
\begin{tabular}{lcccc}
\toprule
Method & Extra & Total & AUROC & Diff. \\
\midrule
\textbf{Anchor + peer} (reference) & 1 & 2 & 0.820 & \\
Pooled greedy SE-2 (anchor, peer) & 1 & 2 & 0.820 & $0.0$ (identical ranking) \\
Same-model, agreement: anchor + 1 sample & 1 & 2 & 0.756 & $+6.5$ $[+5.7, +7.3]$ \\
SE-2 (single model) & 2 & 3 & 0.746 & $+7.4$ $[+6.7, +8.3]$ \\
Pooled SE-2 (1 target + 1 peer) & 2 & 3 & 0.784 & $+3.7$ $[+2.9, +4.4]$ \\
SE-5 (single model) & 5 & 6 & 0.802 & $+1.9$ $[+1.1, +2.7]$ \\
\midrule
\textbf{Anchor + alt + peer} (reference) & 2 & 3 & 0.852 & \\
Pooled greedy SE-3 (anchor, alt, peer) & 2 & 3 & 0.846 & $+0.6$ $[+0.5, +0.7]$ \\
Same-model, agreement: anchor + 2 samples & 2 & 3 & 0.788 & $+6.4$ $[+5.7, +7.1]$ \\
Pooled SE-2 (1 target + 1 peer) & 2 & 3 & 0.784 & $+6.9$ $[+6.1, +7.6]$ \\
SE-3 (single model) & 3 & 4 & 0.775 & $+7.7$ $[+7.0, +8.5]$ \\
Pooled SE-3 (2 target + 1 peer) & 3 & 4 & 0.813 & $+3.9$ $[+3.3, +4.6]$ \\
4-call configuration (alt, peer, peer-alt) & 3 & 4 & 0.869 & $-1.7$ $[-2.0, -1.4]$ \\
Same-model, agreement: anchor + 4 samples & 4 & 5 & 0.814 & $+3.8$ $[+3.2, +4.5]$ \\
Pooled SE-5 (3 target + 2 peer) & 5 & 6 & 0.837 & $+1.5$ $[+0.8, +2.2]$ \\
Same-model, agreement: anchor + 9 samples & 9 & 10 & 0.836 & $+1.7$ $[+1.1, +2.3]$ \\
Pooled SE-10 (5 target + 5 peer) & 10 & 11 & 0.851 & $+0.1$ $[-0.6, +0.7]$ \\
SE-10 (single model) & 10 & 11 & 0.824 & $+2.8$ $[+2.2, +3.5]$ \\
\bottomrule
\end{tabular}
\end{table}

\subsection{Entailment-based clustering}\label{sec:entail}
Both the reconciler and semantic entropy use normalized exact match for short text, so paraphrased answers on TriviaQA and NQ-Open count as different, whereas published semantic entropy clusters by bidirectional entailment~\citep{kuhn2023semantic}. We re-ran both tasks with an entailment relation, applied to every method including the contract's comparison, using a small NLI cross-encoder (Appendix~\ref{app:entail}). It lowered AUROC on the two tasks for every agreement-based method, by 1.1 to 2.6 points, and changed no full-grid comparison by more than 0.4 points (anchor + alt + peer minus SE-10: $+2.3$ under both relations). SE-10 is still not above SE-5 on the same three models, so cluster fragmentation does not explain that pattern. An audit of 200 merged pairs found 8\% clearly wrong merges. We did not test how such merges affect each method, and the effect need not be the same: a wrong merge changes the contract's comparison only when it involves the anchor, but can change the size of several clusters in an entropy over ten samples. We therefore report these results as robustness under this particular NLI model, not as a general property, and keep exact match as the primary relation for every other table.

\subsection{Why other routes help: error coupling}\label{sec:mechanism}
Table~\ref{tab:mechanism} measures the mechanism directly. For each task it gives the accuracy of each check type and, on items where the anchor is wrong, how often the check is also wrong and how often it gives the \emph{same} wrong answer, which is the case the contract cannot detect. A $T{=}0.8$ sample of the same model repeats a wrong anchor 41--77\% of the time. A peer repeats it 19--44\% of the time, and the alternate prompt 27--55\%. On every task the peer is the least coupled check, and on the multiple-choice tasks the temperature sample is the most coupled, by 18 to 36 points. The cost of a less coupled check is lower agreement on correct items: when the anchor is right, the peer agrees with it on 67--97\% of items against 83--98\% for the temperature sample, which is why the contract automates fewer items.

\begin{table}[t]
\centering\small
\caption{Error coupling per task, mean over 8 models. ``Acc.'': accuracy of the check's own answer (a peer's mean accuracy equals the anchor's because the pairs are symmetric). ``Same wrong'': P(check gives the anchor's wrong answer $\mid$ anchor wrong). ``Agree if right'': P(check agrees $\mid$ anchor right).}
\label{tab:mechanism}
\resizebox{\linewidth}{!}{\begin{tabular}{lc|ccc|ccc|ccc}
\toprule
 & Anchor & \multicolumn{3}{c|}{Alternate prompt} & \multicolumn{3}{c|}{Peer} & \multicolumn{3}{c}{Temperature 0.8 (same model)} \\
Task & Acc. & Acc. & Same wrong & Agree if right & Acc. & Same wrong & Agree if right & Acc. & Same wrong & Agree if right \\
\midrule
GSM8K & 0.946 & 0.899 & 0.43 & 0.93 & 0.946 & 0.32 & 0.97 & 0.927 & 0.46 & 0.96 \\
SVAMP & 0.943 & 0.935 & 0.39 & 0.97 & 0.943 & 0.30 & 0.97 & 0.936 & 0.43 & 0.98 \\
MMLU & 0.805 & 0.788 & 0.54 & 0.91 & 0.805 & 0.44 & 0.90 & 0.793 & 0.72 & 0.95 \\
ARC & 0.941 & 0.941 & 0.49 & 0.97 & 0.941 & 0.41 & 0.97 & 0.932 & 0.77 & 0.98 \\
TriviaQA & 0.686 & 0.658 & 0.37 & 0.86 & 0.686 & 0.22 & 0.83 & 0.664 & 0.46 & 0.91 \\
NQ-Open & 0.262 & 0.244 & 0.27 & 0.74 & 0.262 & 0.19 & 0.67 & 0.251 & 0.41 & 0.83 \\
\bottomrule
\end{tabular}}
\end{table}

\paragraph{Alternate-prompt accuracy.} Asking for the equations first lowers accuracy on GSM8K from 0.946 to 0.899 and on SVAMP from 0.943 to 0.935, and the alternate prompt is 0.0 to 2.8 points less accurate on the other tasks. This is a further reason to commit the anchor rather than the check: the alternate prompt is chosen to take a different route, not a better one.

\subsection{Held-out selection and a fresh task}\label{sec:confirm}
We tested several checker combinations on the same data, so we guard against selecting a lucky one in two ways. First, we split every cell 50/50 by a SHA-256 hash of the item identifier, pick the best two-call and three-call recipe on the dev half, and report the test half only (Table~\ref{tab:confirm}). The dev half picks anchor + peer and anchor + alt + peer, the same recipes as on all items. On the test half the three-call recipe is above SE-10 by 1.9 points $[+0.8, +3.1]$; the two-call recipe's differences from SE-5 ($-1.0$ $[-2.3, +0.3]$) and SE-3 ($+1.1$ $[-0.1, +2.3]$) both have intervals containing zero. Against the cross-model baselines, the test half repeats the main grid: the three-call recipe is $+0.3$ $[-0.9, +1.5]$ from pooled SE-3, $0.0$ $[-1.2, +1.5]$ from pooled SE-5 and $-1.6$ $[-2.0, -1.3]$ from the four-call configuration. Intervals in this section use the clustered bootstrap of Section~\ref{sec:setup}, restricted to the test half. On chronology: the alternate prompts, the checker assignments and the three recipe families were fixed before any result was computed; the dev/test split was defined (by hash) after we had seen aggregate results for the full grid, so it guards against picking a lucky recipe among the six, not against earlier design choices made with the aggregate results in view. We therefore treat the held-out half as a post hoc robustness check, and CommonsenseQA, added after every design choice was fixed, as the stronger confirmation.

Second, we ran the full pool (anchor, 10 creative samples, alternate prompt) on CommonsenseQA for all 8 models after these recipes were fixed. The task played no part in designing the prompts, picking the checkers or choosing the recipes. On it the two-call recipe is above SE-5 by 1.8 points $[+0.2, +3.6]$, and the three-call recipe above SE-10 by 4.9 points $[+3.0, +6.8]$ (0.800 against 0.751), on 7 of 8 models. The cross-model baselines were run on the same pool: the three-call recipe is $+2.4$ $[+0.5, +4.2]$ above pooled SE-3, $+1.2$ $[-0.6, +2.9]$ from pooled SE-5 and $-2.2$ $[-2.7, -1.6]$ from the four-call configuration; at 2 calls it trails semantic entropy over the same two greedy answers by 1.5 points $[-2.1, -0.9]$. This is the same ordering as the main grid, with a wider margin at 3 calls. The run cost USD 5.02.

\begin{table}[t]
\centering\small
\caption{Post hoc robustness check and fresh-task confirmation. Left: the test half of the main grid (a split made after aggregate results were seen) ($n=24{,}344$ evaluations), with recipes picked on the dev half. Right: CommonsenseQA, 8 models $\times$ 1,000 items, never used for design. AUROC is the mean over cells; brackets give the paired 95\% interval of the difference in points (clustered bootstrap).}
\label{tab:confirm}
\resizebox{\linewidth}{!}{\begin{tabular}{lcccc}
\toprule
 & \multicolumn{2}{c}{Held-out test half} & \multicolumn{2}{c}{CommonsenseQA} \\
\cmidrule(lr){2-3}\cmidrule(lr){4-5}
Method & AUROC & vs SE & AUROC & vs SE \\
\midrule
Greedy + verbalized confidence & 0.663 & & 0.615 & \\
Anchor + creative (ablation) & 0.750 & & 0.660 & \\
Anchor + alt & 0.783 & & 0.738 & \\
\textbf{Anchor + peer} & 0.811 & SE-5: $-1.0$ $[-2.3, +0.3]$ & 0.752 & SE-5: $+1.8$ $[+0.2, +3.6]$ \\
\textbf{Anchor + alt + peer} & 0.846 & SE-10: $+1.9$ $[+0.8, +3.1]$ & 0.800 & SE-10: $+4.9$ $[+3.0, +6.8]$ \\
 & & pooled SE-3: $+0.3$ $[-0.9, +1.5]$ & & pooled SE-3: $+2.4$ $[+0.5, +4.2]$ \\
 & & pooled SE-5: $0.0$ $[-1.2, +1.5]$ & & pooled SE-5: $+1.2$ $[-0.6, +2.9]$ \\
Semantic entropy, 3 / 5 / 10 & 0.800 / 0.821 / 0.827 & & 0.711 / 0.734 / 0.751 & \\
Pooled SE, 3 / 5 / 10 & 0.843 / 0.846 / 0.845 & & 0.776 / 0.789 / 0.794 & \\
4-call (alt, peer, peer-alt) & 0.862 & & 0.822 & \\
\bottomrule
\end{tabular}}
\end{table}

\subsection{Per-model results and checker strength}\label{sec:permodel}
Figure~\ref{fig:permodel} in Appendix~\ref{app:extra} shows per-model means. The three-call recipe is above SE-10 on all 8 models. The two-call recipe is above SE-5 on 5 of 8 models and below it on the two smallest, Ministral 8B ($-3.9$ points) and GLM 4.7-flash ($-4.9$), and on GLM 5 ($-1.0$). On the small models a peer of the same size makes errors that are partly shared, and their own samples disagree often, which is the regime where semantic entropy is strongest. We read this as suggestive rather than a size law: the pairs also differ in architecture (dense versus mixture-of-experts) and in reasoning mode (gpt-oss).

Against pooled SE-5 the per-model picture is mixed (Table~\ref{tab:permodelfull} and Figure~\ref{fig:permodel}). The three-call contract is above it on Mistral Large 3 ($+1.7$), gpt-oss-120b ($+0.8$) and gpt-oss-20b ($+2.5$), equal on Qwen3 32B ($0.0$), and below it on Ministral 8B ($-1.6$), Qwen3 235B A22B ($-0.6$), GLM 4.7-flash ($-0.9$) and GLM 5 ($-1.4$). The grid mean of $+0.1$ therefore averages per-model differences of up to 2.5 points in either direction; it is not a per-model result. The four-call configuration is the best method on all 8 models.

Averaged over every other-family checker rather than the fixed peer, a weaker checker gives 0.799 and a stronger one 0.843 at two calls. A checker much weaker than the target adds little; the most accurate model, GLM 5, has only weaker checkers available, which is why it benefits least.

\subsection{Operational measurements}\label{sec:ops}

\begin{figure}[t]
\centering
\includegraphics[width=\linewidth]{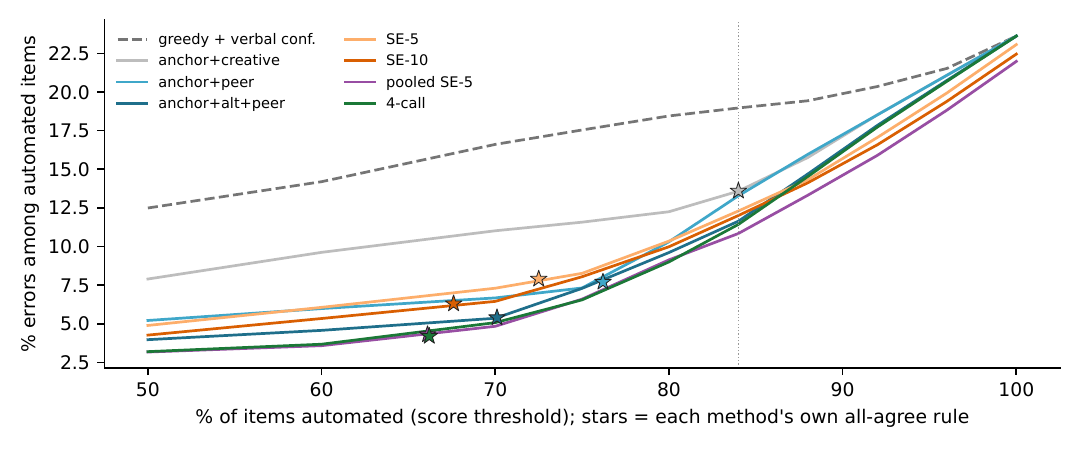}
\caption{Risk--coverage curves on the main grid under the score-threshold policy: error rate among automated items against the share of items automated. Stars mark each method's own all-agree rule (the strict gate for the contract). The dotted line is the 84\% operating point of Table~\ref{tab:gating}.}
\label{fig:riskcov}
\end{figure}

\paragraph{Gating: two policies.}\label{sec:gating} The contract's deployment policy acts on \AGREE{} only. For anchor + alt + peer this \emph{strict gate} automates 70.1\% of items with a 5.4\% error rate among them $[4.9, 5.9]$, against 23.6\% overall, and sends 84.0\% of all errors to review; 16\% of errors are still acted on. Semantic entropy's analogous rule (all samples agree) automates 72.5\% at 7.9\% error with 5 samples and 67.6\% at 6.3\% with 10. These rules automate different shares of items, so Table~\ref{tab:gating} (right) also compares a second, \emph{score-threshold} policy: every method's score is thresholded to automate exactly 84.0\% of items, the share the two-call temperature-based ablation automates under its own rule. For the contract this admits the highest-confidence \DISAGREE{} items, so it is a ranking experiment, not the stated deployment policy. Ties are broken at random with a fixed seed. Under it the three-call contract has an 11.6\% error rate among automated items $[11.0, 12.3]$. Every method is evaluated on the same items, so paired differences (same bootstrap draw for both methods) are much tighter than these marginal intervals. The contract's rate is 0.6 points below SE-5's $[0.4, 0.9]$ and 0.4 below SE-10's $[0.1, 0.6]$; its difference from pooled SE-3 (0.2 points lower, $[-0.1, +0.4]$) is unresolved; it is 0.8 points above pooled SE-5 $[0.6, 1.0]$, whose voted answer is more accurate, and 0.2 above the four-call configuration $[0.1, 0.3]$. Under the strict gate, pooled SE-3 automates the same share as the contract (difference 0.1 points, $[-0.4, +0.3]$) at an error rate 0.2 points higher $[+0.0, +0.4]$, an interval that reaches zero; pooled SE-5 and the four-call configuration automate 4.0 and 3.9 points fewer items at error rates 1.1 and 1.2 points lower (all four intervals exclude zero). At that automation rate 41\% of errors are acted on. Figure~\ref{fig:riskcov} gives the full risk--coverage curves, with each method's own all-agree rule marked. Intervals come from the item-clustered bootstrap of Section~\ref{sec:setup}, which applies each draw of item identifiers to all 8 models. \AGREE{} is evidence that different routes reach the same answer, not proof that the answer is right.

\begin{table}[t]
\centering\small
\caption{Two gating policies on the main grid (overall error rate 23.6\%). Left, strict gate: each method's own rule (act only when all samples agree; for the contract, \AGREE). Right, score-threshold policy: every method's score thresholded to automate 84.0\% of items. ``Caught'': share of all errors sent to review; ``precision'': share of escalated items that are errors. Marginal item-clustered 95\% intervals on the error rates are within $\pm 0.8$ points; paired differences against the three-call contract, which are much tighter, are given in the text. No cell is marked best, because only differences against the contract were tested.}
\label{tab:gating}
\resizebox{\linewidth}{!}{\begin{tabular}{lccc|ccc}
\toprule
 & \multicolumn{3}{c|}{Strict gate (own rule)} & \multicolumn{3}{c}{Score threshold at 84.0\% automation} \\
Method & Automated & Error (auto.) & Caught & Error (auto.) & Caught & Precision \\
\midrule
Greedy + verbalized confidence & -- & -- & -- & 19.0\% & 32.5\% & 48.0\% \\
Anchor + creative (ablation) & 84.0\% & 13.6\% & 51.7\% & 13.6\% & 51.7\% & 76.3\% \\
Anchor + peer & 76.2\% & 7.7\% & 75.2\% & 13.3\% & 52.7\% & 77.9\% \\
Anchor + alt + peer & 70.1\% & 5.4\% & 84.0\% & 11.6\% & 58.6\% & 86.5\% \\
Semantic entropy, 5 samples & 72.5\% & 7.9\% & 75.2\% & 12.3\% & 55.2\% & 79.6\% \\
Semantic entropy, 10 samples & 67.6\% & 6.3\% & 81.0\% & 12.0\% & 55.1\% & 77.3\% \\
\midrule
Pooled SE-3 (2 target + 1 peer) & 70.1\% & 5.6\% & 83.2\% & 11.8\% & 57.4\% & 83.6\% \\
Pooled SE-5 (3 target + 2 peer) & 66.1\% & 4.3\% & 87.1\% & 10.8\% & 58.6\% & 80.4\% \\
4-call (alt, peer, peer-alt) & 66.2\% & 4.2\% & 88.2\% & 11.4\% & 59.3\% & 87.6\% \\
\bottomrule
\end{tabular}}
\end{table}

\paragraph{Stability.}\label{sec:stability} The reconciler is deterministic, but the hosted API is not: it exposes no seed, and repeated greedy calls can differ. We repeated the anchor and alternate-prompt calls for every model on the first 200 items of every cell (9,600 items) and compared the two runs (Figure~\ref{fig:stability} in Appendix~\ref{app:extra}). The anchor itself changes on 7.7\% of items (2.4\% for Ministral 8B to 12.3\% for Qwen3 32B), and the SE-5 answer changes on 12.1\% when five fresh samples are drawn. The gate decision changes on 5.2\% of items for anchor + alt + peer, 5.5\% for anchor + peer, and 11.6\% for the temperature-based check. The repeat run re-drew every call the gate uses, including the peer's answer, so the flip rate includes the peer's own drift. For a like-for-like comparison we also computed semantic entropy's gate flip rate (the all-agree rule of SE-5 on two disjoint sets of five samples): it is 9.9\% (6.0\% for GLM 5 to 17.0\% for GLM 4.7-flash), close to twice the three-call contract's. On the same 9,600 evaluations the paired difference is 4.7 points $[3.9, 5.4]$ (item-clustered bootstrap; contract 5.2\% $[4.7, 5.8]$, SE-5 9.9\% $[9.2, 10.5]$). Figure~\ref{fig:stability} shows committed-answer changes and gate flips as separate bars. So the honest claim is this: given its samples, a decision is exactly reproducible from its record, and a changed equality rule can be re-applied to old records; the samples themselves are not reproducible on this API, and the gate built on greedy checks is about twice as stable as one built on a sampled check.

\paragraph{Latency.} In the fixed, non-adaptive configurations the calls are independent and can run in parallel, so latency is that of the slowest call; the hybrids and format-repair calls add a sequential step. In single-client measurements the median is 0.69\,s for one greedy call, 0.99\,s for anchor + peer and 1.13\,s for anchor + alt + peer; semantic-entropy latency was estimated from greedy-call timings, not measured (Appendix~\ref{app:latency}).

\paragraph{Cost.} All costs are computed from the token counts the API reported for the calls each method actually uses, including format-repair calls (0.2--0.3\% of calls), at the provider's public on-demand list prices. At list prices the three-call contract costs USD 0.208 per 1,000 items in the main grid against 0.354 for SE-5 and 0.708 for SE-10; pooled semantic entropy costs the same as single-model entropy at the same sample count (USD 0.212 / 0.354 / 0.708), and the four-call configuration USD 0.276 (Table~\ref{tab:crossmodel}). ``Equal budget'' in Section~\ref{sec:crossmodel} therefore holds in dollars as well as calls, within 2\%. A cross-family check also means operating a second model endpoint. The cheapest useful recipe, anchor + cheap, costs USD 0.090 and is 0.6 points below SE-3 $[-1.7, +0.5]$, a difference whose sign the interval leaves unresolved.

\section{Deployment Considerations}\label{sec:deploy}
The contract needs, per deployment, a fixed anchor model, a fixed list of checks, a decision key and an equality rule per key type. Reconciliation makes no model call; its work is linear in the number of checks plus the cost of parsing and comparing the fields, which is negligible next to a generation. Because it is pure, a stored record (samples with their sources, key, outcome and confidence) replays any decision. The same property has an operational cost: if the equality rule is later changed, re-applying it to historical records can turn a past \AGREE{} into \DISAGREE. The record then shows that an action was taken on evidence the current rule would not accept. We recommend versioning the equality rule in the record and treating such flips as a review queue rather than silently rewriting history. The total inference cost for this paper, across all runs including development, was about USD 91 at list prices.

\paragraph{Reproducibility.} All samples were generated through AWS Bedrock in the \texttt{us-west-2} region between October and November 2026, with model identifiers \texttt{mistral.ministral-3-8b-instruct}, \texttt{mistral.mistral-large-3-675b-instruct}, \texttt{qwen.qwen3-32b-v1:0}, \texttt{qwen.qwen3-235b-a22b-2507-v1:0}, \texttt{openai.gpt-oss-20b-1:0}, \texttt{openai.gpt-oss-120b-1:0}, \texttt{zai.glm-4.7-flash} and \texttt{zai.glm-5}; the entailment check used \texttt{cross-encoder/nli-deberta-v3-small}. Dataset splits: GSM8K test, SVAMP full, MMLU test, ARC-Challenge test, TriviaQA rc.nocontext validation, NQ-Open validation, CommonsenseQA validation, each subsampled to 1,000 items with seed 0.

\section{Limitations and Ethics}\label{sec:limits}
\begin{itemize}
\item \AGREE{} is evidence that different routes reach the same answer, not of correctness. At its own rule the three-call contract still acts on 16\% of errors, and errors that every route shares are hard to detect this way (Table~\ref{tab:mechanism}).
\item Cross-family checks need a second model of comparable strength. A much weaker checker adds little, and on the two smallest models the two-call contract trails 5-sample semantic entropy.
\item Checker answers come from the shared pool: each model's anchor for the same item. The repeat run bounds what a fresh call changes for the gate, but every number other than stability uses one draw per call.
\item The alternate prompts were written once for three task types. Other tasks need their own, and a poorly chosen alternate prompt could add correlated errors.
\item All models were served by one provider in one region. Latency figures are single-client, single-request and load-dependent; semantic-entropy latency is estimated from greedy-call timings, not measured (Appendix~\ref{app:latency}).
\item The structured and date key types are specified and unit-tested but not evaluated; every reported result uses a numeric, categorical or short-text key.
\item The decision key must be declared by the pipeline author. Normalization-based equality counts paraphrased free-text answers as disagreements; our entailment check used a small NLI model (DeBERTa-v3-small) rather than the DeBERTa-large of the original semantic-entropy work, and a stronger model could cluster differently.
\item As a complete pipeline the contract does not beat cross-model baselines at equal budget: its differences from pooled semantic entropy at 3 and 5 samples have intervals containing zero, it trails entropy over the same two answers by 0.9 points at 2 calls, and it trails its own four-call configuration by 1.7 points. Per model, pooled SE-5 is ahead of it on four of the eight models. As a detector of the anchor's errors it is ahead of pooled semantic entropy with 3 and 5 samples (the baseline's budget; the contract uses three calls) under both tie-breaker conventions, and with 10 samples the point estimates differ by less than one point, with a sign that depends on the convention (Tables~\ref{tab:anchorgraded} and~\ref{tab:controlled}). Its case rests on the fixed committed answer, the explicit outcome, the bounded confidence and the small specification, none of which is impossible to add to a baseline.
\item No equivalence margin was pre-specified, and the dev/test split was defined after aggregate results had been seen (Section~\ref{sec:confirm}).
\item The hybrid in the ordering of \citet{xue2025verify} uses our fixed peer and all-agree rule, not their verifier prompt or threshold, so it is an adaptation.
\item The perturbation and cross-model ideas are not new~\citep{zhang2023sac3,xue2025verify}; this paper contributes a minimal deterministic configuration and its measurement, not the checks themselves.
\item Escalation moves work to people. The review queue must be staffed, or the gate becomes a silent drop.
\end{itemize}
All datasets are public benchmarks; no personal or user data was collected. The models were accessed through a hosted inference API under their published licenses.

\section{Conclusion}
Keeping the greedy answer and checking it against greedy answers that take a different route, through another prompt or another model family, gives a decision contract that is pure, replayable and cheap. Where the calls come from matters more than how many there are, and this holds for the baselines too: semantic entropy over samples from two models beats semantic entropy over ten samples from one. Under matched scoring the three-call contract beats single-model semantic entropy up to 10 samples at under a third of its cost. Against cross-model semantic entropy, the pipelines' point estimates at 3 and 5 samples differ by less than one point, with intervals containing zero that reach 1.2 points; as a detector of the anchor's errors the contract is ahead of that baseline with 3 and 5 samples, and with 10 the point estimates differ by less than one point, with a sign that depends on the tie-breaker. A fourth check adds 1.7 points. The measured error coupling explains the ordering: a model's own temperature samples repeat its wrong answers far more often than another prompt or another model does. What the contract offers over appropriately instrumented baselines is not better detection but a small, fully specified configuration: a committed answer that sampling cannot change, an explicit outcome instead of a threshold, a confidence bounded by the anchor's, and a record that replays every decision.

\bibliography{main}
\bibliographystyle{tmlr}

\appendix
\section{Alternate prompts}\label{app:prompts}
The original prompts ask for two lines, \texttt{Answer:} and \texttt{Confidence: <0-100>}. The alternate prompts keep that format and change the route:
\begin{itemize}
\item \textbf{MMLU, ARC, CommonsenseQA}: ``Pick the correct option for the question below.'' The options are listed in reverse order and relabeled; the returned letter is mapped back to the original labeling. A round-trip check on all items found no mapping errors.
\item \textbf{GSM8K, SVAMP}: ``Read the word problem, write the equation or equations that describe it, then compute the result. Keep it brief.''
\item \textbf{TriviaQA, NQ-Open}: ``Answer the trivia question. On the first line, name in a few words the key fact you are relying on.'' Forty extra output tokens are allowed for that line.
\end{itemize}
The templates were written once, before any alternate-prompt result was seen, and not revised.

\section{Per-model results}\label{app:permodel}

\begin{table}[h]
\centering\small
\caption{Per-model AUROC (mean over 6 tasks), main grid. ``Pooled SE-5'': 3 target + 2 peer samples; ``4-call'': alt, peer and peer-alt checks.}
\label{tab:permodelfull}
\resizebox{\linewidth}{!}{\begin{tabular}{lccccccccc}
\toprule
Model & creative & alt & peer & alt + peer & 4-call & SE-3 & SE-5 & SE-10 & Pooled SE-5 \\
\midrule
Ministral 8B & 0.729 & 0.759 & 0.806 & 0.854 & 0.877 & 0.830 & 0.845 & 0.843 & 0.870 \\
Mistral Large 3 & 0.773 & 0.804 & 0.823 & 0.863 & 0.873 & 0.795 & 0.812 & 0.805 & 0.846 \\
Qwen3 32B & 0.734 & 0.787 & 0.816 & 0.852 & 0.867 & 0.794 & 0.815 & 0.830 & 0.852 \\
Qwen3 235B A22B & 0.707 & 0.768 & 0.809 & 0.837 & 0.853 & 0.768 & 0.785 & 0.796 & 0.843 \\
gpt-oss-20b & 0.804 & 0.802 & 0.857 & 0.871 & 0.889 & 0.842 & 0.856 & 0.850 & 0.846 \\
gpt-oss-120b & 0.798 & 0.816 & 0.849 & 0.859 & 0.871 & 0.824 & 0.826 & 0.838 & 0.850 \\
GLM 4.7-flash & 0.755 & 0.785 & 0.802 & 0.856 & 0.874 & 0.840 & 0.851 & 0.853 & 0.865 \\
GLM 5 & 0.734 & 0.778 & 0.788 & 0.826 & 0.847 & 0.772 & 0.798 & 0.815 & 0.839 \\
\bottomrule
\end{tabular}}
\end{table}

\begin{table}[h]
\centering\small
\caption{Peer pairs: committing either member. Accuracy is that member's greedy accuracy; AUROC is for the pair with that member as anchor. Cost is the same for both directions.}
\label{tab:pairs}
\resizebox{\linewidth}{!}{\begin{tabular}{llcccc}
\toprule
Pair & Anchor & Accuracy & AUROC (peer) & AUROC (alt + peer) & USD / 1k items \\
\midrule
Ministral 8B + GLM 4.7-flash & Ministral 8B & 0.706 & 0.806 & 0.854 & 0.040 \\
 & GLM 4.7-flash & 0.730 & 0.802 & 0.856 & 0.040 \\
Qwen3 32B + gpt-oss-20b & Qwen3 32B & 0.746 & 0.816 & 0.852 & 0.072 \\
 & gpt-oss-20b & 0.730 & 0.857 & 0.871 & 0.072 \\
Mistral Large 3 + GLM 5 & Mistral Large 3 & 0.805 & 0.823 & 0.863 & 0.331 \\
 & GLM 5 & 0.817 & 0.788 & 0.826 & 0.331 \\
Qwen3 235B + gpt-oss-120b & Qwen3 235B & 0.790 & 0.809 & 0.837 & 0.120 \\
 & gpt-oss-120b & 0.787 & 0.849 & 0.859 & 0.120 \\
\bottomrule
\end{tabular}}
\end{table}

\paragraph{Outcome and parse rates.} For the two-call temperature-based ablation, outcomes over the main grid are 84.0\% \AGREE, 15.6\% \DISAGREE, 0.20\% \SF, 0.12\% \SFs{} and 0.05\% \INVALID{} (summing to 100.0\% before rounding). The gpt-oss models, which received extra output tokens, have \INVALID, \SF{} and \SFs{} rates at or below 0.07\%, in line with the other models, so the token allowance did not favor them.

\section{Additional figures and per-model notes}\label{app:extra}

\begin{figure}[h]
\centering
\includegraphics[width=\linewidth]{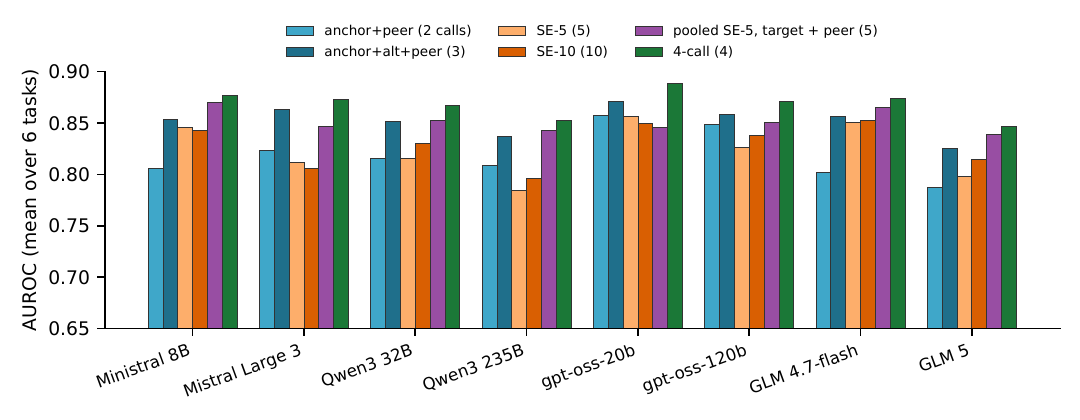}
\caption{Per-model AUROC (mean over 6 tasks) for the two- and three-call contract, the four-call configuration, 5- and 10-sample semantic entropy and pooled 5-sample semantic entropy.}
\label{fig:permodel}
\end{figure}

\begin{figure}[h]
\centering
\includegraphics[width=\linewidth]{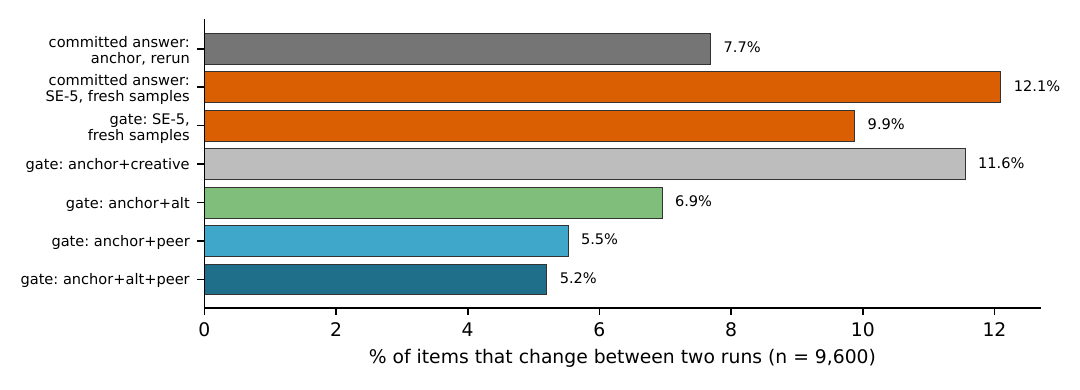}
\caption{Share of items that change between two runs of the same calls (first 200 items of each of the 48 cells). The top two bars are changes in the committed answer; the rest are gate flips (automate versus escalate). For SE-5 the two runs are disjoint sets of five $T{=}0.8$ samples; for the contract every call is repeated, including the peer's.}
\label{fig:stability}
\end{figure}

\paragraph{Which answer to commit.} If a peer is available, why not commit the peer's answer? Within each peer pair the two members are close in accuracy (differences of 0.3 to 2.4 points) and cost the same as a pair, so the choice is symmetric: committing the more accurate member gains 0.3--2.4 accuracy points and loses 0.4--4.1 points of AUROC, because its errors are harder for the weaker member to catch (Table~\ref{tab:pairs} in Appendix~\ref{app:permodel}). The contract supports either choice; it requires only that the anchor be fixed per deployment, so that the record states whose answer was committed.

\section{Entailment-based clustering: full results}\label{app:entail}
\paragraph{Setup.} Two answers are equivalent if they match exactly or if an NLI model judges that each entails the other given the question. We used a DeBERTa-v3-small cross-encoder trained on MNLI and SNLI (\texttt{cross-encoder/nli-deberta-v3-small}), smaller than the DeBERTa-large model of \citet{kuhn2023semantic}, because the larger model was too slow for the 297,463 answer pairs on our hardware. Grading still uses the gold aliases. Table~\ref{tab:entail} gives every method under both relations. The relation applies to TriviaQA and NQ-Open; the other four tasks are unchanged, so full-grid means move by at most one third of the short-text change (SE-5 from 0.824 to 0.808 on the two tasks, the three-call contract from 0.837 to 0.820).

\paragraph{Audit of merges.} We drew a random sample of 200 of the 9,812 pairs the NLI model merged and labeled each as acceptable, debatable or clearly wrong. A single rater labeled them; no inter-rater agreement was measured. 175 were acceptable (paraphrases, reorderings, added qualifiers such as ``yale'' and ``yale university''), 9 debatable and 16 clearly wrong (8\%), 10 of these pairs of distinct short TriviaQA proper nouns (``bagpuss'' and ``casper'', ``kiwano'' and ``kiwi'') that the small model treats as mutually entailing. A larger NLI model would likely remove most of these; we did not measure how each method responds to them.

\begin{table}[h]
\centering\small
\caption{Mean AUROC with normalized exact match (``exact'') and with bidirectional entailment on the two short-text tasks (``entail''). Left: TriviaQA and NQ-Open only (16 cells). Right: full grid (48 cells).}
\label{tab:entail}
\resizebox{\linewidth}{!}{\begin{tabular}{lcccc}
\toprule
 & \multicolumn{2}{c}{Short-text tasks} & \multicolumn{2}{c}{Full grid} \\
\cmidrule(lr){2-3}\cmidrule(lr){4-5}
Method & Exact & Entail & Exact & Entail \\
\midrule
Greedy + verbalized confidence & 0.662 & 0.662 & 0.663 & 0.663 \\
Anchor + creative (ablation) & 0.752 & 0.741 & 0.754 & 0.750 \\
Anchor + alt & 0.782 & 0.769 & 0.787 & 0.783 \\
Anchor + peer & 0.802 & 0.787 & 0.819 & 0.814 \\
Anchor + alt + peer & 0.837 & 0.820 & 0.852 & 0.847 \\
SE-3 / SE-5 / SE-10 & 0.815 / 0.824 / 0.822 & 0.800 / 0.808 / 0.807 & 0.808 / 0.824 / 0.829 & 0.803 / 0.818 / 0.824 \\
Pooled SE-3 / 5 / 10 & 0.845 / 0.838 / 0.828 & 0.824 / 0.815 / 0.802 & 0.849 / 0.852 / 0.846 & 0.841 / 0.844 / 0.837 \\
4-call (alt, peer, peer-alt) & 0.852 & 0.836 & 0.869 & 0.863 \\
\bottomrule
\end{tabular}}
\end{table}

\paragraph{Native greedy AUROC.} Without its tie-breaker the greedy baseline's score is constant for every valid anchor and $-1$ for an invalid one, so its native AUROC would be exactly 0.5 if all anchors were valid. The reported 0.515 comes from the 86 invalid anchors (0.18\% of evaluations), all of which are graded wrong and ranked lowest.

\section{Latency measurements}\label{app:latency}
We timed 50 MMLU and 50 GSM8K items per model with one request in flight, for the anchor and the alternate prompt. In the fixed, non-adaptive configurations measured here the calls are independent and can run in parallel, so latency is that of the slowest call; the hybrids and format-repair calls add a sequential step that these figures do not include. Averaged over models, the median is 0.69\,s for one greedy call, 0.99\,s for anchor + peer and 1.13\,s for anchor + alt + peer. We did not time sampled calls, so the semantic-entropy figures are \emph{estimates}: for each item we drew 5 or 10 latencies from the same model's measured greedy calls and took the maximum, giving about 1.45\,s and 1.78\,s. Run one after another, the medians are 1.54\,s and 2.22\,s for the contract against an estimated 4.4\,s and 8.9\,s for SE-5 and SE-10. These are single-region, single-client figures with one request in flight. Concurrent production latency differs because of queuing, rate limits and shared capacity, and we did not measure it.

\section{Design history}\label{app:history}
We record how the design and analysis changed, so that readers can judge selection effects. The first design checked the anchor against one temperature-0.8 sample of the same model. During revision we noted that our comparison gave that score a verbalized-confidence tie-breaker the semantic-entropy baseline did not get; under matched scoring the design trails SE-5 by 7 points, and we now report it as an ablation (anchor + creative). The alternate prompt and the peer were then added, and the checker assignment fixed, before the CommonsenseQA run. Later revisions added the cross-model baselines, the entailment re-run, the controlled comparison of Table~\ref{tab:controlled}, the anchor-graded view and the clustered bootstrap. An exploratory structured-extraction run on WikiBio was dropped from this version because its automatically derived gold labels could not be validated by annotators blind to the model outputs; no reported result used it. The held-out split of Section~\ref{sec:confirm} was defined after aggregate results had been seen. Two code defects found during revision were corrected without changing a reported conclusion. First, the multi-check reconciler in the accompanying code (\texttt{reconcile\_n}) dropped invalid samples before deciding, so an invalid anchor whose checks agreed returned \AGREE; it now follows Section~\ref{sec:contract} exactly and is tested against every combination of values and validity for $m\le 3$. No reported number used it: every multi-check score and gate in the paper is computed by a separate helper that requires all samples to be valid, and on the main grid the corrected function's \AGREE{} coincides with that helper's gate on all 48,000 items, while the old function differed on 150 (0.31\%). Second, one script drew its bootstrap independently per task--model cell rather than once per task; its intervals were recomputed with the shared item-level draw, which moved their bounds by at most 0.5 points and changed one interval the paper reports (anchor + cheap against SE-3, now $[-1.7, +0.5]$).

\end{document}